\documentclass[10pt,conference,letterpaper]{IEEEtran}
\usepackage{times}
\usepackage{amsmath}
\usepackage{graphicx}
\graphicspath{ {images/} }
\usepackage[caption=false,font=footnotesize]{subfig}
\usepackage{booktabs}
\usepackage{multirow}
\usepackage{tabularx}
\usepackage{tabu}
\usepackage{amsthm}
\usepackage{bm}
\usepackage{xspace}
\usepackage{pgfplots}
\usepackage{enumerate}
\usepackage{algorithm}
\usepackage{algorithmic}
\usepackage{balance}
\pgfplotsset{compat=1.13}
\usetikzlibrary{arrows,automata,positioning}
\title{NP-GLM:\\A Non-Parametric Method for Temporal\\Link Prediction}
\author{%
{Sina Sajadmanesh{\small $~^{\#1}$}, Jiawei Zhang{\small $~^{*2}$}, Hamid R. Rabiee{\small $~^{\#3}$} }%
\vspace{1.6mm}\\
\fontsize{10}{10}\selectfont\itshape
$^{\#}$\,Department of Computer Engineering, Sharif University of Technology\\
Tehran, Iran\\
\fontsize{9}{9}\selectfont\ttfamily\upshape
$^{1}$\,sajadmanesh@ce.sharif.edu\\
$^{3}$\,rabiee@sharif.edu%
\vspace{1.2mm}\\
\fontsize{10}{10}\selectfont\rmfamily\itshape
$^{*}$\,Department of Computer Science, University of Illinois at Chicago\\
Chicago, IL, USA\\
\fontsize{9}{9}\selectfont\ttfamily\upshape
$^{2}$\,jzhan9@uic.edu
}
\begin{document}
\maketitle
\newtheorem{definition}{Definition}
\newcommand{\descr}[1]{\smallskip\noindent\textbf{#1}}
\newcommand{\npglm}{{\textsc{Np-Glm}}\xspace}

\begin{abstract}
In this paper, we try to solve the problem of temporal link prediction in information networks. This implies predicting the time it takes for a link to appear in the future, given its features that have been extracted at the current network snapshot. To this end, we introduce a probabilistic non-parametric approach, called \emph{Non-Parametric Generalized Linear Model} (\npglm), which infers the hidden underlying probability distribution of the link advent time given its features. We then present a learning algorithm for \npglm and an inference method to answer time-related queries. Extensive experiments conducted on both synthetic data and real-world Sina Weibo social network demonstrate the effectiveness of \npglm in solving temporal link prediction problem vis-\`a-vis competitive baselines.
\end{abstract}
\section{Introduction}\label{sec:intro}
Link prediction is the problem of prognosticating a certain relationship, like interaction or collaboration, between two entities in a networked system that are not connected already \cite{lu2011link}. This problem has attracted a considerable attention and has found its application in various interdisciplinary domains, such as viral marketing, bioinformatics, recommender systems, and social network analysis \cite{wasserman1994social}. For example, suggesting new friends in an online social network \cite{liben2007link} or predicting drug-target interactions in a biological network \cite{chen2012drug} are two quite different tasks that both rely on link prediction.

The problem of link prediction has a long literature and is studied extensively. In recent years, newer studies have shifted from traditional link prediction toward new domains, such as time-aware link prediction \cite{dhote2013survey}, link prediction in heterogeneous networks \cite{shi2017survey}, and multi-network link prediction \cite{kivela2014multilayer}. Most of these works have ultimately formulated the link prediction problem as a binary classification task, i.e. predicting \textbf{whether} a link will appear in the network in the future. However, an interesting problem, which we call it \emph{temporal link prediction} in this paper, could be predicting \textbf{when} a link will emerge between two entities in the network. Examples of this problem includes predicting the time that two individuals become friends in a social network, or the time that two authors collaborate on writing a paper \cite{sun2012will}. Inferring the link formation time in advance can be very useful in many concrete applications. For example, if a social network recommender system could predict the relationship time between two people, then it can suggest a friendship close to that time since it has a relatively higher chance to be accepted.

%More formally, the goal of temporal link prediction studied in this paper is to predict by when a link will appear between two nodes, given the state and the characteristics of the network and its topological features up to the current point of time.
%
%, e.g. predicting the time that two individuals become friends in a social network, or the time that two authors collaborate on writing a paper \cite{sun2012will}.
%The goal of temporal link prediction studied in this paper is to answer such time-related queries about future links. 

The temporal link prediction is a challenging problem which cannot be solve trivially for three main reasons. First, the formulation of temporal link prediction is quite different from traditional binary link prediction due to the involvement of time and the necessity of considering network evolution time-line. As opposed to the works concerning the binary link prediction, there are very little works on temporal link prediction that aim to answer the ``when'' question. Second, we only know the creation time of links that are already present at the network and for those links that are yet to happen, which are excessive in number versus the existing ones, we lack such information. Finally, a common approach to this problem is to infer a probability distribution over time for each pair of nodes given their features, and answer time-related queries about the link creation time between the two nodes using the inferred distribution. In this case, the underlying distribution of the link's time is unknown and considering any specific distribution as a priori may be far from reality or limit the solution generality.

In this paper, we propose a probabilistic non-parametric approach to solve the problem of temporal link prediction and address its challenges. To this end, we first define the temporal link prediction problem formally and formulate the approach to solve it generally. Next, we present \emph{Non-Parametric Generalized Linear Model} (\npglm) which models the distribution of link creation time given its feature vector. The strength of this non-parametric model is that it is capable of learning the underlying distribution of the data as well as the amount of contribution of each extracted feature for the link advent time in the network. Inferring such probability distribution, we propose an inference method to answer queries, like the most probable time by which a link will appear between two nodes, or the probability of link creation between two nodes during a specific period. Extensive experiments on both synthetic dataset and real-world social network data demonstrate that the proposed method works well in predicting the link's apparition time versus the relevant ones.

The rest of this paper is organized as follows. In Section \ref{sec:problem}, we provide introductory concepts and formally define the problem of temporal link prediction. Next, we introduce our proposed \npglm method in Section \ref{sec:method}, explaining its learning method and how to answer inference queries. Experimental results are described in Section \ref{sec:results}. Section \ref{sec:related} discusses related works and finally in Section \ref{sec:conclusion}, we conclude the paper.

\section{Problem Formulation}\label{sec:problem}
In this section, we formulate the temporal link prediction problem and introduce some important concepts and definitions used throughout the paper.

%\begin{table*}
%%\renewcommand{\arraystretch}{2}
%\centering
%\caption{Characteristics of Some Probability Distributions Used for Event-Time Modeling}
%\label{table:dists}
%\begin{tabu} to \textwidth {X X[c] X[c] X[c] X[c]}
%\toprule
%Distribution & Density function & Reliability function & Intensity function & Cumulative intensity\\
%& $f_T(t)$ & $S(t)$ & $\lambda(t)$ & $\Lambda(t)$\\[1pt]
%\midrule % In-table horizontal line
%Exponential & $\alpha\exp(-\alpha t)$ & $\exp(-\alpha t)$ & $\alpha$ & $\alpha t$\\[4pt]
%%\midrule
%Rayleigh & $\frac{t}{\sigma^2}\exp(-\frac{t^2}{2\sigma^2})$ & $\exp(-\frac{t^2}{2\sigma^2})$ & $\frac{t}{\sigma^2}$ & $\frac{t^2}{2\sigma^2}$\\[4pt]
%%\midrule % In-table horizontal line
%Gompertz & $\alpha e^t\exp\left\lbrace -\alpha(e^t-1) \right\rbrace$ & $\exp\left\lbrace -\alpha(e^t-1) \right\rbrace$ & $\alpha e^t$ & $\alpha e^t$\\[2pt]
%%\midrule % In-table horizontal line
%Power-Law & $\frac{\alpha\beta^\alpha}{t^{\alpha+1}}$ & $\left(\frac{\beta}{t}\right)^\alpha$ & $\frac{\alpha}{t}$ & $\alpha\ln(t)$\\
%\bottomrule % Bottom horizontal line
%\end{tabu}
%\end{table*}

\subsection{Temporal Link Prediction}
The aim of this paper is to predict the time of link creation in social networks.
Formally, given the feature vector $x_l$ for a missing link $l$ extracted in time $t_0$, we want to predict $t_l$, which shows how long after $t_0$ the link $l$ will appear in the network. A probabilistic approach to this problem is to model the conditional distribution $f_T(t_l\mid x_l)$.

\subsection{Data Description}
Suppose that we have a snapshot of the network at the time $t_0$, and we have seen the evolution of the network (the emergence of new links) in the time interval $[t_0,t_e]$ called \textit{time window}. Based on the existence state of the links prior to $t_0$, between $t_0$ and $t_e$, and after $t_e$, we can classify links in the following categories:

\begin{enumerate}
\item Links that are already present at time $t_0$.
\item Links that do not exist at $t_0$, but will appear during the time window.
\item Links that remain missing all the time when we reach $t_e$.
\end{enumerate}

Those links that fall within the 2nd and the 3rd categories form our data samples and will be used in the learning procedure. For these links, we extract their feature vectors at time $t_0$. For a link $l$ of the 2nd category, we have seen that it is created at a time like $t_c\in[t_0,t_e]$. So we set $t_l=t_c-t_0$ as the time it takes for the link $l$ to appear after $t_0$, and $y_l=1$ which indicates that we have \emph{observed} its exact creation time. If $l$ is of the 3rd category, we haven't seen its exact creation time, but we know it is definitely after $t_e$. For such samples, which we call the \emph{censored} ones, we set $t_l=t_e-t_f$ and $y_l=0$ to indicate that the recorded time is in fact less than the real one. These type of links are also of interest because their features will give us some information about their time falling after $t_e$. As a result, each link $l$ is associated with a triple $(x_l,y_l,t_l)$ representing its feature vector, its observation status, and the time it takes to appear, respectively. In Section \ref{sec:method}, we propose \npglm which is a supervised method to relate $x_l$ to $t_l$ by estimating $f_T(t_l\mid x_l)$ in a non-parametric fashion.

\subsection{Basic Concepts}
Here we define some essential concepts that are necessary to study before we proceed to the proposed method. Generally, the formation of a link between two nodes in the network can simply be considered as an event with its occurring time as a random variable $T$ coming from a density function $f_T(t)$. Regarding this, we can have the following definitions:

\begin{definition}[Survival Function]
Given the density $f_T(t)$, the survival function denoted by $S(t)$, is the probability that an event occurs after a certain value of $t$, which means:
\begin{equation}
    S(t) = P(T > t) = \int_t^\infty f_T(t)dt
\end{equation}
\end{definition}

\begin{definition}[Intensity Function]
The intensity function (or failure rate function), denoted by $\lambda(t)$, is the instantaneous rate of occurring the event at any time $t$ given the fact that the event has not occurred yet:
\begin{equation}
    \lambda(t)=\lim_{\Delta t\rightarrow 0}\frac{P(t\le T\le t+\Delta t\mid T\ge t)}{\Delta t}
\end{equation}
\end{definition}

%\begin{definition}[Cumulative Intensity Function]
%The cumulative intensity function, denoted by $\Lambda(t)$, is the area under the intensity function up to a point $t$:
%\begin{equation}
%    \Lambda(t)=\int_0^t\lambda(t)dt
%\end{equation}
%\end{definition}

The relations between the density, survival, and intensity functions come directly from their definitions as follows:

\begin{equation}\label{eq:intensity}
    \lambda(t)=\frac{f_T(t)}{S(t)}
\end{equation}
\begin{equation}\label{eq:reliability}
    S(t)=\exp(-\int_0^t\lambda(t)dt)
\end{equation}
%\begin{equation}\label{eq:density}
%    f_T(t)=\lambda(t)\exp(-\Lambda(t))
%\end{equation}

%Table \ref{table:dists} shows the density, reliability, intensity, and cumulative intensity functions of some widely-used distributions for event time modeling.
\section{Proposed Method}\label{sec:method}
In this section we introduce our proposed model, called \emph{Non-Parametric Generalized Linear Model}, to solve the problem of temporal link prediction. As we talked about in previous sections, we are going to model the distribution $f_T(t\mid x)$ so that we can answer the time-related queries using the feature vector $x$ for a missing link in the network. The recent approach \cite{sun2012will} has considered a specific distribution for $t$ (e.g. Exponential distribution) and then related $x$ to $t$ using a Generalized Linear Model. The major drawback of this approach is that we need to know the exact distribution of time, or at least, we could guess the best one that fits. The alternative way that we follow is to learn the shape of $f_T(t\mid~x)$ from the data using a non-parametric solution.

\subsection{Model Description}
Looking at the Eq.~\ref{eq:intensity}, we see that the density function can be specified uniquely with its intensity function. Since the intensity function often has a simpler form than the density itself, if we learn the shape of the intensity function, then we can infer the entire distribution eventually. Therefore, we focus on learning the shape of the conditional intensity function $\lambda(t\mid x)$ from the data, and then accordingly infer the conditional density function $f_T(t\mid x)$ based on the learned intensity.
In order to reduce the hypothesis space of the problem and avoid the curse of dimensionality, we assume that $\lambda(t\mid x)$, which is a function of both $t$ and $x$, can be factorized into two separate positive functions as the following:
\begin{equation}\label{eq:lambda}
\lambda(t\mid x)=g(w^Tx)h(t)
\end{equation}
where $g$ is a functions of $x$ which captures the effect of features via a linear transformation using coefficient vector $w$ independent of $t$, and $h$ is a function of $t$ which captures the effect of time independent from $x$. This assumption, referred to as proportional hazards condition \cite{breslow1975analysis}, holds in GLM formulations of many event-time modeling distributions, such as Exponential, Rayleigh, Power-Law, and so on. Our goal is now to fix the function $g$ and learn the coefficient vector $w$ and the function $h$ from the training data. We begin by the likelihood function of the data which is as follows:

\begin{equation}
\prod_{i=1}^{N}f_T(t_i\mid x_i)^{y_i}P(T\ge t_i\mid x_i)^{1-y_i}\\
\end{equation}
The likelihood consists of the product of two parts: The first part is the contribution of those samples for which we have seen their exact formation time in terms of their density function. The second part on the other hand, is the contribution of the censored samples. For these samples, we use the probability of the formation time being greater than the recorded one. Applying the Eq.~\ref{eq:intensity}, \ref{eq:reliability}, and \ref{eq:lambda}, the likelihood function becomes:
\begin{equation}
\prod_{i=1}^{N}\left[g(w^Tx_i)h(t_i)\right]^{y_i}\exp\lbrace-g(w^Tx_i)\int_{0}^{t_i}h(t)dt\rbrace
\end{equation}

Since we don't know the form of $h(t)$, we cannot directly calculate the integral appeared in the likelihood function. To deal with this problem, we approximate $h(t)$ with a piecewise constant function that changes just in $t_i$ s. Therefore, the integral over $h(t)$, denoted by $H(t)$, becomes a series:
\begin{equation}\label{eq:cumh}
H(t_i)=\int_{0}^{t_i}h(t)dt \simeq \sum_{j=1}^{i}h(t_j)(t_j-t_{j-1})
\end{equation}
assuming samples are sorted by $t$ in increasing order, without loss of generality. The function $H(t)$ defined above plays an important role in both learning and inference phases. In fact, both the learning and inference phases rely on $H(t)$ instead of $h(t)$, which we will see later in this paper.
Replacing the above series in the likelihood, we end up with the following log-likelihood function:

\begin{equation}\label{eq:logl}
\begin{split}
\log\mathcal{L}
=\sum_{i=1}^{N}\Big\lbrace& y_i\left[\log g(w^Tx_i) + \log h(t_i)\right]\\&-g(w^Tx_i)\sum_{j=1}^{i}h(t_j)(t_j-t_{j-1})\Big\rbrace\\
\end{split}
\end{equation}

The log-likelihood function depends on the vector $w$ and the function $h(t)$. In the next part, we explain an iterative learning algorithm to learn both $w$ and $h$ collectively.

\subsection{Model Learning}
Maximizing the log-likelihood function (Eq.~\ref{eq:logl}) rely on the choice of the function $g$. There are no particular limits on the choice of $g$ except that it must be a non-negative function. For example, both quadratic and exponential functions of $w^Tx$ will do the trick. Here, we proceed with $g(w^Tx)=\exp(w^Tx)$ since it yields a convex optimization function with respect to $w$. Subsequent equations can be derived for other choices of $g$ in the same way.

Setting the log-likelihood derivative with respect to $h(t_k)$ to zero, yields a closed form solution for $h(t_k)$:
\begin{equation}
h(t_k)=\frac{y_k}{(t_k-t_{k-1})\sum_{i=k}^{N}\exp(w^Tx_i)}
\end{equation}

Applying Eq.~\ref{eq:cumh}, we get the following for $H(t_i)$:
\begin{equation}
H(t_i)=\sum_{j=1}^{i}\frac{y_j}{\sum_{k=j}^{N}\exp(w^Tx_k)}
\end{equation}
which depends on the vector $w$. On the other hand, we cannot obtain a closed form solution for $w$ from the log-likelihood function. Therefore, we turn to use Gradient-based optimization methods to find the optimal value of $w$. The negative log-likelihood function with respect to $w$, denoted by $NL(w)$ is as follows:

\begin{equation}\label{eq:nlw}
NL(w)=\sum_{i=1}^{N}\left\lbrace\exp(w^Tx_i)H(t_i)-y_iw^Tx_i\right\rbrace
\end{equation}
which depends on the function $H$. As the learning of both $w$ and $H$ depends on each other, they should be learned collectively. Here, we use an iterative algorithm to learn $w$ and $H$ alternatively. We begin with a random vector $w^{(0)}$. Then in each iteration $\tau$, we first update $H^{(\tau)}(t_i)$ via Eq.~\ref{eq:cumh} using $w^{(\tau-1)}$. Second, we optimize Eq.~\ref{eq:nlw} using the values of $H^{(\tau)}(t_i)$ to obtain $w^{(\tau)}$. We continue this procedure until convergence.

\subsection{Model Inference}
In this part, we come across answering the common inference queries after learning the vector $\hat{w}$ and the function $\hat{H}$ estimated using $N$ training samples. For a test link $l$ with feature vector $x_l$, the following queries can be answered:\\

\descr{Ranged Probability.} What is the probability for the link $l$ to be formed between time $t_\alpha$ and $t_\beta$? This is equivalent to calculating $P(t_\alpha \le T \le t_\beta \mid x_l)$, which by definition is equal to:
\begin{equation}\label{eq:ranged}
\begin{split}
P(t_\alpha\le T \le t_\beta \mid x_l) = S(t_\alpha\mid x_l) - S(t_\beta\mid x_l)\\
= \exp\{-g(\hat{w}^Tx_l)\hat{H}(t_\alpha)\} - \exp\{-g(\hat{w}^Tx_l)\hat{H}(t_\beta)\}
\end{split}
\end{equation}
The problem here is to obtain the values of $\hat{H}(t_\alpha)$ and $\hat{H}(t_\beta)$, as $t_\alpha$ and $t_\beta$ may not be among $t_i$s of the training samples, for which $\hat{H}(.)$ is estimated. To calculate $\hat{H}(t_\alpha)$, we find $k\in\{1,2,\dots,N\}$ such that $t_k\le t_\alpha < t_{k+1}$. Due to the piecewise constant assumption for $h(.)$, we get:
\begin{equation}\label{eq:inf1}
\hat{h}(t_\alpha)=\frac{\hat{H}(t_\alpha)-\hat{H}(t_k)}{t_\alpha-t_k}
\end{equation} 
On the other hand, since $h(.)$ only changes in $t_i$s, we have:
\begin{equation}\label{eq:inf2}
\hat{h}(t_\alpha)=\hat{h}(t_{k+1})=\frac{\hat{H}(t_{k+1})-\hat{H}(t_k)}{t_{k+1}-t_k}
\end{equation}
Combining Eq.~\ref{eq:inf1} and \ref{eq:inf2}, we have:
\begin{equation}\label{eq:inf3}
\hat{H}(t_\alpha)=\hat{H}(t_k)+(t_\alpha-t_k)\frac{\hat{H}(t_{k+1})-\hat{H}(t_k)}{t_{k+1}-t_k}
\end{equation}
Following the similar approach, we can calculate $\hat{H}(t_\beta)$, and then answer the query using Eq.~\ref{eq:ranged}. The dominating operation here is to find the value of $k$. Since we have $t_i$s sorted beforehand, this operation can be done using a binary search with $O(\log N)$ time complexity.\\

\descr{Quantile.} By how long the link $l$ will be formed with probability $\alpha$? This question is equivalent to find the time $t_\alpha$ such that $P(T \le t_\alpha\mid x_l)=\alpha$. By definition, we have:
\begin{equation*}
\begin{split}
1-P(T \le t_\alpha\mid x_l)=S(t_\alpha\mid x_l)&=\exp\{-g(\hat{w}^Tx_l)\hat{H}(t_\alpha)\}\\
&=1-\alpha
\end{split}
\end{equation*}
Taking logarithm of both sides and rearranging, we get:
\begin{equation}\label{eq:inf4}
\hat{H}(t_\alpha)=-\frac{\log(1-\alpha)}{g(\hat{w}^Tx_l)}
\end{equation}
To find $t_\alpha$, we first find $k$ such that $\hat{H}(t_k)\le\hat{H}(t_\alpha)<\hat{H}(t_{k+1})$. We eventually have $t_k\le t_\alpha < t_{k+1}$ since $H(.)$ is a non-decreasing function due to the function $h$ being non-negative. Therefore, we again end up with Eq.~\ref{eq:inf3}, which by rearranging we get:
\begin{equation}\label{eq:inf5}
t_\alpha=(t_{k+1}-t_k)\frac{\hat{H}(t_\alpha)-\hat{H}(t_k)}{\hat{H}(t_{k+1})-\hat{H}(t_k)}+t_k
\end{equation}
By combining the Eq.~\ref{eq:inf4} and \ref{eq:inf5}, we can obtain the value of $t_\alpha$ which is the answer to the quantile query. It worth mentioning that if $\alpha=0.5$ then $t_\alpha$ becomes the median of the distribution $f_T(t\mid x_l)$. Here again the dominant operation is to find the value of $k$, which due to the non-decreasing property of $\hat{H}(.)$ can be found using a binary search with $O(\log N)$ time complexity.
\section{Experiments}\label{sec:results}

We conduct extensive experiments on both synthetic and real-world datasets to evaluate the effectiveness of \npglm.

\subsection{Experiments on synthetic data}
We use synthetic data to verify the correctness of \npglm and its learning algorithm. Since \npglm is a non-parametric method, we generate synthetic data using various parametric models with previously known random parameters, and evaluate how can \npglm learn the parameters and the underlying distribution of the generated data.

\descr{Experiment Setup.}
We consider generalized linear models of two widely used distributions for event-time modeling, Rayleigh and Gompertz, as the ground truth models for generating the synthetic data. To generate a total of $N$ data samples with $d$-dimensional feature vectors, consisting $N_o$ non-censored (observed) samples and remaining $N_c=N-N_o$ censored ones, we use the following procedure:

\begin{enumerate}
\item Draw a weight vector $w\sim\mathcal{N}(0,I_d)$, where $I_d$ is the $d$-dimensional identity matrix.
\item Draw scalar intercept $b\sim\mathcal{N}(0,1)$.
\item For $i=1\dots N$ do 
\begin{enumerate}[i]
\item Draw feature vector $x_i\sim\mathcal{N}(0,I_d)$.
\item Set distribution parameter $\alpha_i=\exp(w^Tx_i+b)$
\item Draw time $t_i$ based on the distribution:
\begin{itemize}
    \item[] Rayleigh: $t_i\sim\alpha_i~t\exp\{-0.5\alpha_it^2\}$.
    \item[] Gompertz: $t_i\sim\alpha_i~e^t\exp\{-\alpha_i(e^t-1)\}$.
\end{itemize}
\end{enumerate}
\item Sort pairs $(x_i,t_i)$ by $t_i$ in ascending order.
\item For $i=1\dots~N_o$ set $y_i=1$
\item For $i=(N_o+1)\dots~N$ set $y_i=0$
\end{enumerate}

For all synthetic experiments, we generate 10-dimensional feature vectors ($d=10$) and set $g(w^Tx)=\exp(w^Tx)$.
We repeat every experiment 100 times and report the average.

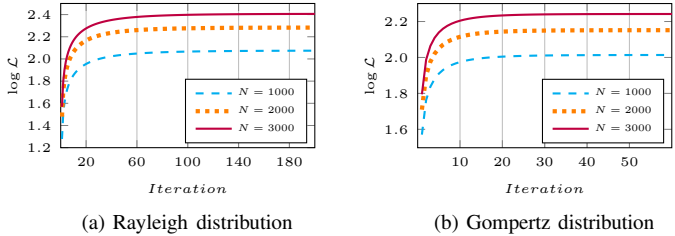
\begin{figure}[t]
\hfill
    \subfloat[Rayleigh distribution]{
    \begin{tikzpicture}[trim axis left, trim axis right]
\begin{axis}
[
tiny,
width=0.56\columnwidth,
height=3.5cm,
legend pos=south east,
legend style={font=\tiny,nodes={scale=0.75, transform shape}},
xmajorgrids,
y tick label style={
    /pgf/number format/.cd,
        fixed,
        fixed zerofill,
        precision=1,
    /tikz/.cd
},
xlabel=$Iteration$,
ylabel=$\log\mathcal{L}$,
ylabel shift = -4 pt,
ymax=2.5,
ymin=1.2,
xmin=0,
xmax=200,
xtick={20,60,...,180},
restrict x to domain=0:200,
legend entries={${\tiny N=1000}$, $N=2000$, $N=3000$},
]
\addplot[color=cyan,  thick, dashed] table{results/cvg_ray_1000.txt};
\addplot[color=orange,ultra thick, dotted] table{results/cvg_ray_2000.txt};
\addplot[color=purple,thick] table{results/cvg_ray_3000.txt};
\end{axis}
\end{tikzpicture}
    }\hspace{1cm}    
    \subfloat[Gompertz distribution]{
   \begin{tikzpicture}[trim axis left, trim axis right]
\begin{axis}
[
tiny,
width=0.56\columnwidth,
height=3.5cm,
legend pos=south east,
legend style={font=\tiny,nodes={scale=0.75, transform shape}},
xmajorgrids,
y tick label style={
    /pgf/number format/.cd,
        fixed,
        fixed zerofill,
        precision=1,
    /tikz/.cd
},
xlabel=$Iteration$,
ylabel=$\log\mathcal{L}$,
ylabel shift = -4 pt,
ymax=2.3,
xmin=0,
xmax=60,
xtick={10,20,...,50},
restrict x to domain=0:100,
legend entries={$N=1000$, $N=2000$, $N=3000$},
]
\addplot[color=cyan  ,thick, dashed] table{results/cvg_gom_1000.txt};
\addplot[color=orange,ultra thick, dotted] table{results/cvg_gom_2000.txt};
\addplot[color=purple,thick] table{results/cvg_gom_3000.txt};
\end{axis}
\end{tikzpicture}
    }
    \caption{Convergence of \npglm's average log-likelihood ($\log\mathcal{L}$) for different number of training samples ($N$). Censoring ratio has been set to 0.5.}
    \label{fig:syn-cvg-n}
\end{figure}

\descr{Experiment Results.}
As \npglm's learning is done in an iterative manner, we first analyzed whether this algorithm converges as the number of iterations increase. We recorded the log-likelihood of \npglm, averaged over the number of training samples $N$ in each iteration. We repeated this experiments for $N\in\{1000,2000,3000\}$ with a fixed censoring ratio of 0.5, which means half of the samples are censored. The result is depicted in Fig.~\ref{fig:syn-cvg-n}. We can see that the algorithm successfully converges with a rate depending on the underlying distribution. For the Rayleigh distribution, it requires about 100 iterations to converge but for Gompertz, this reduces to about 30. Also, we see that using more training data results in achieving more log-likelihood as expected.

\begin{figure}[t]
\hfill
    \subfloat[Rayleigh distribution]{
    \begin{tikzpicture}[trim axis left, trim axis right]
\begin{axis}
[
tiny,
width=0.56\columnwidth,
height=3.5cm,
legend pos=south east,
legend style={font=\tiny,nodes={scale=0.75, transform shape}},
xmajorgrids,
y tick label style={
    /pgf/number format/.cd,
        fixed,
        fixed zerofill,
        precision=1,
    /tikz/.cd
},
xlabel=$Iteration$,
ylabel=$\log\mathcal{L}$,
ylabel shift = -8 pt,
ymin=-2,
xmin=0,
xmax=100,
xtick={10,30,...,90},
restrict x to domain=0:200,
legend entries={5\% censoring, 25\% censoring, 50\% censoring},
]
\addplot[color=cyan  ,thick, dashed] table{results/cvg_ray_5.txt};
\addplot[color=orange,ultra thick, dotted] table{results/cvg_ray_25.txt};
\addplot[color=purple,thick] table{results/cvg_ray_50.txt};
\end{axis}
\end{tikzpicture}
    }\hspace{1cm}    
    \subfloat[Gompertz distribution]{
   \begin{tikzpicture}[trim axis left, trim axis right]
\begin{axis}
[
tiny,
width=0.56\columnwidth,
height=3.5cm,
legend pos=south east,
legend style={font=\tiny,nodes={scale=0.75, transform shape}},
xmajorgrids,
y tick label style={
    /pgf/number format/.cd,
        fixed,
        fixed zerofill,
        precision=1,
    /tikz/.cd
},
xlabel=$Iteration$,
ylabel=$\log\mathcal{L}$,
ylabel shift = -4 pt,
xmin=0,
xmax=60,
xtick={10,20,...,50},
restrict x to domain=0:60,
legend entries={5\% censoring, 25\% censoring, 50\% censoring},
]
\addplot[color=cyan  ,thick, dashed] table{results/cvg_gom_5.txt};
\addplot[color=orange,ultra thick, dotted] table{results/cvg_gom_25.txt};
\addplot[color=purple,thick] table{results/cvg_gom_50.txt};
\end{axis}
\end{tikzpicture}
    }
    \caption{Convergence of \npglm's average log-likelihood ($\log\mathcal{L}$) for different censoring ratios with 1K samples.}
    \label{fig:syn-cvg-c}
\end{figure}
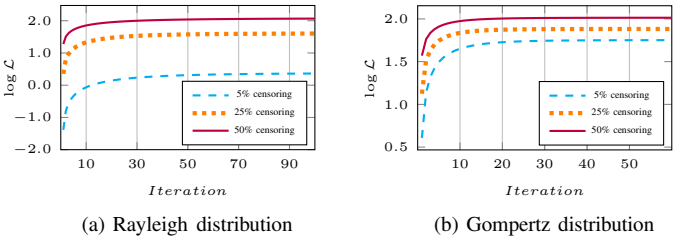

In Fig.~\ref{fig:syn-cvg-c}, we fixed $N=1000$ and performed the same experiment this time using different censoring ratios. According to the figure, we see that by increasing the censoring ratio, the convergence rate increases. This is because \npglm infers the values of $H(t)$ for all $t$ in the time window. Therefore, as the censoring ratio increases, the time window is decreased, so \npglm has to infer a fewer number of parameters, leading to a faster convergence. Note that as opposed to Fig.~\ref{fig:syn-cvg-n}, here a higher log-likelihood doesn't necessarily indicate a better fit, due to the likelihood marginalization we get by the censored samples.

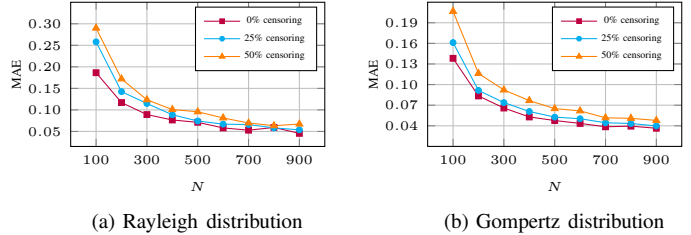
\begin{figure}[t]
\hfill  
    \subfloat[Rayleigh distribution]{
    \begin{tikzpicture}[trim axis left, trim axis right]
\begin{axis}
[
tiny,
width=0.56\columnwidth,
height=3.5cm,
legend pos=north east,
legend style={font=\tiny,nodes={scale=0.75, transform shape}},
grid,
y tick label style={
    /pgf/number format/.cd,
        fixed,
        fixed zerofill,
        precision=2,
    /tikz/.cd
},
xlabel=$ N $,
ylabel=MAE,
ylabel shift = -4 pt,
ymax=0.35,
xmin=0,
xmax=1000,
ytick={0.05,0.10,...,0.35},
xtick={100,300,...,900},
restrict x to domain=0:900,
legend entries={0\% censoring, 25\% censoring, 50\% censoring},
]
\addplot[color=purple,mark=square*,mark size=1.1,] table{results/mae_ray.txt};
\addplot[color=cyan,mark=*,mark size=1.1,] table{results/mae_ray_25.txt};
\addplot[color=orange,mark=triangle*,mark size=1.5,] table{results/mae_ray_50.txt};
\end{axis}
\end{tikzpicture}
    }\hspace{1cm}
    \subfloat[Gompertz distribution]{
    \begin{tikzpicture}[trim axis left, trim axis right]
\begin{axis}
[
tiny,
width=0.56\columnwidth,
height=3.5cm,
legend pos=north east,
legend style={font=\tiny,nodes={scale=0.75, transform shape}},
grid,
y tick label style={
    /pgf/number format/.cd,
        fixed,
        fixed zerofill,
        precision=2,
    /tikz/.cd
},
xlabel=$ N $,
ylabel=MAE,
ylabel shift = -4 pt,
ymax=0.22,
ymin=0.01,
xmin=0,
xmax=1000,
xtick={100,300,...,900},
restrict x to domain=0:900,
ytick={0.04,0.07,...,0.21},
legend entries={0\% censoring, 25\% censoring, 50\% censoring},
]
\addplot[color=purple,mark=square*,mark size=1.1,] table{results/mae_gom.txt};
\addplot[color=cyan,mark=*,mark size=1.1,] table{results/mae_gom_25.txt};
\addplot[color=orange,mark=triangle*,mark size=1.5,] table{results/mae_gom_50.txt};
\end{axis}
\end{tikzpicture}
    }
    \caption{\npglm's mean absolute error (MAE) vs the number of training samples ($N$) for different censoring ratios.}
    \label{fig:syn-mae-n}
\end{figure}

Next, we evaluated how good \npglm can infer the parameters used to generate the synthetic data. To this end, we varied the number of training samples $N$ and measured the mean absolute error (MAE) between the learned weight vector $\hat{w}$ and the ground truth. Fig.~\ref{fig:syn-mae-n} illustrates the result for different censoring ratios. It can be seen that as the number of training samples increases, the MAE gradually decreases. The other point to notice is that more censoring ratio results in higher error due to the information loss we get by censoring.

\begin{figure}[t]
\hfill
    \subfloat[Rayleigh distribution]{
    \begin{tikzpicture}[trim axis left, trim axis right]
\begin{axis}
[
tiny,
width=0.56\columnwidth,
height=3.5cm,
legend pos=north east,
legend style={font=\tiny,nodes={scale=0.75, transform shape}},
grid,
y tick label style={
    /pgf/number format/.cd,
        fixed,
        fixed zerofill,
        precision=2,
    /tikz/.cd
},
xlabel=$N_c$,
ylabel=MAE,
ylabel shift = -4 pt,
ymax=0.2,
ymin=0.06,
ytick={0.08,0.10,...,0.2},
xtick={0,40,...,200},
legend entries={$N_o=200$, $N_o=300$, $N_o=400$},
]
\addplot[color=cyan,mark=*,mark size=1.1,] table{results/mae_ray_200.txt};
\addplot[color=orange,mark=triangle*,mark size=1.5,] table{results/mae_ray_300.txt};
\addplot[color=purple,mark=square*,mark size=1.1,] table{results/mae_ray_400.txt};
\end{axis}
\end{tikzpicture}
    }\hspace{1cm}    
    \subfloat[Gompertz distribution]{
   \begin{tikzpicture}[trim axis left, trim axis right]
\begin{axis}
[
tiny,
width=0.56\columnwidth,
height=3.5cm,
legend pos=north east,
legend style={font=\tiny,nodes={scale=0.75, transform shape}},
grid,
y tick label style={
    /pgf/number format/.cd,
        fixed,
        fixed zerofill,
        precision=2,
    /tikz/.cd
},
xlabel=$N_c$,
ylabel=MAE,
ylabel shift = -4 pt,
ymax=0.24,
ymin=0.03,
ytick={0.06,0.09,...,0.21},
xtick={0,40,...,200},
legend entries={$N_o=200$, $N_o=300$, $N_o=400$},
]
\addplot[color=cyan,mark=*,mark size=1.1,] table{results/mae_gom_200.txt};
\addplot[color=orange,mark=triangle*,mark size=1.5,] table{results/mae_gom_300.txt};
\addplot[color=purple,mark=square*,mark size=1.1,] table{results/mae_gom_400.txt};
\end{axis}
\end{tikzpicture}
    }
    \caption{\npglm's mean absolute error (MAE) vs the number of censored samples ($N_c$) for different number of observed samples ($N_o$).}
    \label{fig:syn-mae-c}
\end{figure}
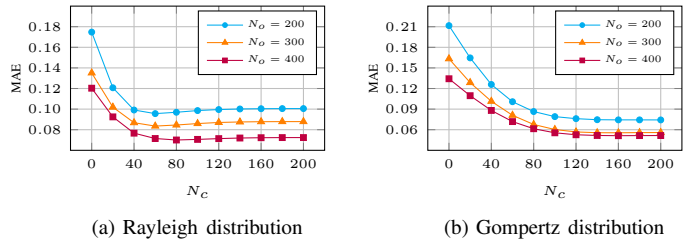

Finally, we investigated whether censored samples are informative or not. For this purpose, we fixed the number of observed samples $N_o$ and changed the number of censored samples from 0 to 200. We measure the MAE between $\hat{w}$ and the ground truth for $N_o\in\{200,300,400\}$. The result is shown in Fig.~\ref{fig:syn-mae-c}. It clearly demonstrates that adding more of censored samples causes the MAE to dwindle up to an extent, after which we get no substantial improvement. This threshold is depended on the underlying distribution. In this case, for Rayleigh and Gompertz it is about 80 and 120, respectively.

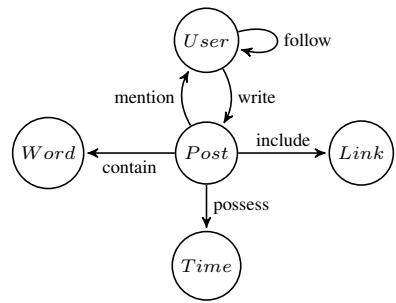
\begin{figure}
\centering
\scriptsize
\begin{tikzpicture}[->,>=stealth',shorten >=1pt,auto,node distance=1.5cm,semithick]

  \node[state] (P)                    {$Post$};
  \node[state] (U) [above       of=P] {$User$};
  \node[state] (W) [left=1.2cm  of P] {$Word$};
  \node[state] (L) [right=1.2cm of P] {$Link$};
  \node[state] (T) [below       of=P] {$Time$};

  \path (U) edge [loop right] node {follow}  (U)
            edge [bend  left] node {write}   (P)
        (P) edge [bend  left] node {mention} (U)
            edge              node {include} (L)
            edge              node {contain} (W)
            edge              node {possess} (T);
\end{tikzpicture}
\caption{Schema of the Weibo network.}
\label{fig:schema}
\end{figure}

\begin{table}
\centering
\caption{Properties of the Weibo Sub-Network}
\label{table:stat}
\scriptsize
\begin{tabu} to \columnwidth {X[l] X[r] c X[l] X[r]}
\toprule
\multicolumn{2}{c}{\# Nodes} & & \multicolumn{2}{c}{\# Relations}\\
\cmidrule(l){1-2} \cmidrule{4-5}
User      & 3,000     & & follow        & 56,441 \\
Post      & 28,900    & & mention       & 6,662  \\
Word      & 1,177,343 & & contain       & 926,033\\
Link (URL)       & 7,524     & & include       & 7,521  \\
Time      & 24    & & write/possess & 28,900 \\
\bottomrule
\end{tabu}
\end{table}

\begin{table}
\centering
\caption{Similarity Meta-Paths Used for Feature Extraction}
\label{table:meta}
\scriptsize
\begin{tabu} to \columnwidth {X[c] X[l]}
\toprule
Meta-Path & Semantic Meaning \\
\midrule
$U\rightarrow~U\leftarrow~U$ & Common followee\\
$U\leftarrow~U\rightarrow~U$ & Common follower\\
$U\rightarrow~P\rightarrow~U\leftarrow~P\leftarrow~U$ & Common mentioned user\\
$U\rightarrow~P\rightarrow~W\leftarrow~P\leftarrow~U$ & Common word in posts\\
$U\rightarrow~P\rightarrow~L\leftarrow~P\leftarrow~U$ & Common referenced URL\\
$U\rightarrow~P\rightarrow~T\leftarrow~P\leftarrow~U$ & Common posting time\\
\bottomrule
\end{tabu}
\end{table}

\subsection{Experiments on real data}
We apply \npglm on real-world dataset to evaluate its effectiveness and compare its performance in predicting the time of link creation vis-\`a-vis different parametric models. 

\descr{Dataset.} We use a dynamic real-world dataset from \emph{Sina Weibo}, which is a Chinese microblogging social network. This dataset, provided by \cite{zhang2013}, is a heterogeneous social network whose meta structure is shown in Fig.~\ref{fig:schema}. It is composed of a static and a dynamic part: The static part describes the overal state of the network at the very first timestamp $t_0=$~September 27th 2012; and the dynamic part reflects the new following links along with their times occurred in a time window of 32 days, between September 28th to October 29th, 2012. Since the original dataset is too massive to process (having about 2 million users and 400 million following links), we confine the number of users to 3000 via random edge sampling with graph induction; a network sampling method which is shown to well preserve the topological structure of the original network \cite{ahmed2013}. The demographic statistics of the sampled network is presented in Table~\ref{table:stat}.

\begin{table*}[t]
\centering
\caption{Performance of Different Methods On the Weibo Dataset Under Different Measures}
\label{table:results}
%\tiny
\begin{tabu} to \textwidth {X[l] X[c] X[c] c X[c] X[c] X[c]}
\toprule
%& 
& \multicolumn{2}{c}{Median Prediction Error} & & \multicolumn{3}{c}{Confidence Interval Prediction Accuracy (\%)}\\
%\cmidrule(l){3-4} \cmidrule{6-8}
\cmidrule(l){2-3} \cmidrule{5-7}
%Dataset & 
Model & MAE & MRE & & 25\%-75\% & 20\%-80\% & 15\%-85\%\\
\midrule
%\multirow{5}{*}{Sub-Network \#1} & 
\npglm & $\bm{10.59\pm0.18}$ & $\bm{4.41\pm0.50}$ & & $\bm{73.83\pm0.48}$ & $\bm{78.98\pm0.45}$ & $\bm{84.27\pm0.51}$\\
%& 
\textsc{Exp-Glm} & $13.12\pm0.09$ & $6.79\pm0.05$ & & $47.78\pm0.19$ & $49.28\pm0.18$ & $51.03\pm0.24$\\
%& 
\textsc{Ray-Glm} & $14.44\pm0.07$ & $9.51\pm0.07$ & & $48.84\pm0.10$ & $49.16\pm0.08$ & $49.55\pm0.07$\\
%& 
\textsc{Pow-Glm} & $12.17\pm0.10$ & $5.77\pm0.11$ & & $51.19\pm0.31$ & $54.83\pm0.30$ & $61.83\pm0.51$\\
%& 
\textsc{Gom-Glm} & $16.18\pm0.02$ & $11.07\pm0.04$ & & $43.73\pm0.17$ & $45.58\pm0.21$ & $46.66\pm0.15$\\
%\midrule
%\multirow{5}{*}{Sub-Network \#2} & 
%\npglm & $\bm{10.19\pm0.10}$ & $\bm{4.43\pm0.07}$ & & $\bm{73.71\pm0.56}$ & $\bm{79.05\pm0.49}$ & $\bm{84.19\pm0.40}$\\
%& \textsc{Exp-Glm} & $12.82\pm0.04$ & $6.67\pm0.05$ & & $47.88\pm0.22$ & $49.41\pm0.26$ & $51.48\pm0.27$\\
%& \textsc{Ray-Glm} & $14.25\pm0.03$ & $9.38\pm0.04$ & & $48.84\pm0.11$ & $49.23\pm0.11$ & $49.67\pm0.12$\\
%& \textsc{Pow-Glm} & $11.65\pm0.06$ & $5.50\pm0.08$ & & $52.07\pm0.25$ & $55.70\pm0.27$ & $62.51\pm0.35$\\
%& \textsc{Gom-Glm} & $16.15\pm0.02$ & $11.03\pm0.05$ & & $44.05\pm0.23$ & $45.75\pm0.19$ & $46.67\pm0.15$\\
%\midrule
%\multirow{5}{*}{Sub-Network \#3} & 
%\npglm & $\bm{10.34\pm0.12}$ & $\bm{4.41\pm0.02}$ & & $\bm{73.77\pm0.31}$ & $\bm{79.10\pm0.32}$ & $\bm{84.34\pm0.36}$\\
%& \textsc{Exp-Glm} & $13.01\pm0.06$ & $6.73\pm0.02$ & & $47.66\pm0.17$ & $49.21\pm0.14$ & $51.02\pm0.13$\\
%& \textsc{Ray-Glm} & $14.37\pm0.03$ & $9.45\pm0.03$ & & $48.81\pm0.08$ & $49.17\pm0.07$ & $49.58\pm0.08$\\
%& \textsc{Pow-Glm} & $11.78\pm0.06$ & $5.53\pm0.06$ & & $51.64\pm0.17$ & $55.19\pm0.24$ & $62.15\pm0.22$\\
%& \textsc{Gom-Glm} & $16.15\pm0.02$ & $11.04\pm0.02$ & & $43.88\pm0.25$ & $45.55\pm0.16$ & $46.55\pm0.16$\\
\bottomrule
\end{tabu}
\end{table*}

\descr{Experiment Setup.}
All pairs of users that establish a following relationship in the time window form our non-censored samples, whose number is about 57,000 pairs. We then randomly pick an equal number of user pairs who do not establish a following relationship, neither at the very first timestamp nor in the time window, as censored ones.

As the Weibo dataset is a heterogeneous social network, we use \emph{meta-paths} \cite{sun2011pathsim} to extract features for each pair of users. Regarding the network schema shown in Fig.~\ref{fig:schema}, we consider the symmetric similarity meta-paths presented in Table~\ref{table:meta}, where User, Post, Link, Time, and Word node types are denoted by $ U $, $ P $, $ L $, $ T $, and $ W $, respectively. For each sample pair of users, we apply the \emph{Path-Count} measure \cite{sun2011pathsim} on each meta-path to obtain a unique feature vector. Due to having different scales for different meta-paths, we normalize the obtained features using z-score.

To challenge the performance of the \npglm, we use ordinary generalized linear models with Exponential, Rayleigh, Power-Law, and Gompertz distributions, denoted as \textsc{Exp-Glm}, \textsc{Ray-Glm}, \textsc{Pow-Glm}, and \textsc{Gom-Glm}, respectively. We use 10-fold cross-validation and report the average results for all the experiments in this section.

\descr{Experiment Results.}
We evaluated the prediction performance of different methods using different sets of measures. First, for each test sample $x_{test}$, we considered the median of the distribution $f_T(t\mid~x_{test})$ as the predicted time for that sample and then compared it to the ground truth time $t_{test}$. Mean absolute error (MAE) and mean relative error (MRE) are used to measure the accuracy of the predicted values. Second, we inferred three different confidence intervals for each test sample and checked whether the ground truth time falls within these confidence intervals or not. Thereby, for each confidence interval, we calculated the percentage of the test samples for which their true times belong to that interval.

Table~\ref{table:results} presents the results obtained for each model under different settings. We can see that our \npglm method performs better than other ones under all measures. Comparing to its closest competitor, \textsc{Pow-Glm}, our method has reduced both MAE and MRE by about 13\% and 23\%, respectively. For confidence interval prediction, \npglm has gained a considerable accuracy in all three cases, which is far better than the other's. In 25\%-75\% confidence interval, \npglm has improved the accuracy by about 44\% relative to \textsc{Pow-Glm}. Under 20\%-80\% confidence interval, again an improvement of about 44\% has been achieved. Finally in 15\%-85\% confidence interval, \npglm can improve the accuracy of  \textsc{Pow-Glm} by about 36\%. These results confirms that the \npglm can better cope with the hidden underlying distribution of link creation time given the link features and it can well estimate this distribution and utilize it to do predictions.

\section{Related Works}\label{sec:related}

%The problem of link prediction has been studied extensively in recent years and many approaches have been proposed to solve the problem \cite{2015arXiv151101868L,wang2015link,wang2014review}.

Previous works on time-aware link prediction have mostly considered temporality in analyzing the long-term network trend over time \cite{dhote2013survey}. Authors in \cite{potgieter2009temporality} have shown that temporal metrics are an extremely valuable new contribution to link prediction, and should be used in future applications. \cite{tylenda2009towards} incorporated temporal information available on evolving social networks for link prediction tasks and proposed a novel node-centric approach to the evaluation of link prediction. \cite{dunlavy2011temporal} focused on the problem of periodic temporal link prediction. They considered bipartite graphs that evolved over time and also considered weighted matrix that contained multilayer data and tensor-based methods for predicting future links. \cite{oyama2011cross} solved the problem of cross-temporal link prediction, in which the links among nodes in different time frames are inferred. they mapped data objects in different time frames into a common low-dimensional latent feature space, and identified the links on the basis of the distance between the data objects. \cite{ozcan2016temporal} proposed a novel link prediction method for evolving networks based on NARX neural network. They take the correlation between the quasi-local similarity measures and temporal evolutions of link occurrences information into account by using NARX for multivariate time series forecasting. \cite{yu2017temporally} developed a novel temporal matrix factorization model to explicitly represent the network as a function of time. They provided results for link prediction as an specific example and showed that their model performs better than the state-of-the-art techniques.

Most of the above works answered the question of \emph{whether} a link will appear in the network. To the best of our knowledge, the only work that has focused on the \emph{when} problem, have proposed a generalized linear model based framework to model the time of link creation \cite{sun2012will}. They consider the building time of links as independent random variables coming from a pre-specified distribution and model the expectation as a function of a linear predictor of the extracted topological features. A shortcoming of this model is that we need to exactly specify the underlying distribution of times. We came over this problem by learning the distribution from the data using a non-parametric solution.
\section{Conclusion}\label{sec:conclusion}
In this paper, we studied the problem of temporal link prediction and proposed a probabilistic non-parametric method, \npglm, to predict the time of link creation in social networks. Our method does not impose any significant assumption on the underlying distribution of the link's advent time given its features, but tries to infer it from the data via a non-parametric approach. Extensive experiments conducted on both synthetic dataset and real-world data from Weibo social network demonstrated the correctness of our method and its effectiveness in predicting the formation time of links.\\

\bibliographystyle{IEEEtran}
\balance
\bibliography{IEEEabrv,IEEEexample}

\end{document}